\pdfoutput=1

\documentclass[11pt]{article}

\usepackage{ACL2023}

\usepackage{times}
\usepackage{latexsym}

\usepackage[T1]{fontenc}

\usepackage[utf8]{inputenc}

\usepackage{microtype}

\usepackage{inconsolata}

\usepackage{graphicx}
\usepackage{algorithm}
\usepackage{algorithmic}
\usepackage{arydshln}
\usepackage{comment}
\usepackage{multirow}
\usepackage{booktabs}
\usepackage{subfigure}

\usepackage{xcolor}

\definecolor{ultramarine}{RGB}{0,32,96}
\definecolor{wrongultramarine}{RGB}{0.07, 0.04, 0.56}
\definecolor{danshuhong}{RGB}{234, 107, 102}
\definecolor{purpleforus}{RGB}{181, 115, 157}
\definecolor{yalv}{RGB}{150, 194, 78}
\definecolor{molv}{RGB}{103, 171, 159}
\definecolor{orangelight}{RGB}{255, 179, 102}
\definecolor{yellowdark}{RGB}{239,236,0}
\definecolor{bluedark}{RGB}{126,166,224}
\definecolor{newpurple}{RGB}{181,115,157}
\definecolor{lightpink}{RGB}{255,208,215}
\definecolor{lightblue}{RGB}{164, 221, 237}

\title{PULSAR: Pre-training with Extracted Healthcare Terms for Summarising Patients’ Problems and Data Augmentation with Black-box Large Language Models}

 \author{Hao Li$^\clubsuit$\footnotemark[1]  
 \footnotemark[2] ,  Yuping Wu$^\clubsuit$\footnotemark[1] , Viktor Schlegel$^{\diamondsuit \clubsuit}$, Riza Batista-Navarro$^\clubsuit$\\ {\bf Thanh-Tung Nguyen$^{\diamondsuit}$} {\bf Abhinav Ramesh Kashyap$^{\diamondsuit \heartsuit}$}, {\bf Xiaojun Zeng$^\clubsuit$} \\ {\bf Daniel Beck$^\spadesuit$}, {\bf Stefan Winkler$^{\diamondsuit \heartsuit}$} \and {\bf Goran Nenadic$^{\clubsuit}$} \\
         $^\clubsuit$ University of Manchester, United Kingdom  
         $^\spadesuit$University of Melbourne, Australia \\
         $^\diamondsuit$ASUS Intelligent Cloud Services (AICS), Singapore\\
         $^\heartsuit$National University of Singapore, Singapore\\
         }

\begin{document}
\maketitle
\renewcommand{\thefootnote}{\fnsymbol{footnote}} 
\footnotetext[1]{These authors contributed equally to this work}
\footnotetext[2]{Corresponding email: \texttt{hao.li-2@manchester.ac.uk}}
\footnotetext[3]{Our code is avaliable at https://github.com/yuping-wu/PULSAR}
\begin{abstract}
Medical progress notes play a crucial role in documenting a patient's hospital journey, including his or her condition, treatment plan, and any updates for healthcare providers. Automatic summarisation of a patient's problems in the form of a ``problem list'' can aid stakeholders in understanding a patient's condition, reducing workload and cognitive bias. 
BioNLP 2023 Shared Task 1A focuses on generating a list of diagnoses and problems from the provider's progress notes during hospitalisation. In this paper, we introduce our proposed approach to this task, which integrates two complementary components \footnotemark[3]. One component employs large language models (LLMs) for data augmentation; the other is an abstractive summarisation LLM with a novel pre-training objective for generating the patients' problems summarised as a list. Our approach was ranked second among all submissions to the shared task.  The performance of our model on the development and test datasets shows that our approach is more robust on unknown data, with an improvement of up to 3.1 points over the same size of the larger model.
\end{abstract}

\section{Introduction}

Medical progress notes are used to document a patient's course in a hospital, including their current condition, treatment plan, and any updates to the plan \citep{DBLP:journals/csr/LiPGVWNRLZCTKR22}. Automated identification of treated  problems from the assessment sections of a progress note in form of a ``problem list'' can help healthcare stakeholders to gain an accurate understanding of the patient's condition, reducing workload and cognitive bias \cite{DBLP:conf/coling/GaoDMXCA22}. This problem list is then used to outline and pursue a detailed treatment plan.  

The majority of studies on clinical summarisation have focused on clinical notes; radiology reports \citep{DBLP:conf/acl-louhi/ZhangDQML18, DBLP:conf/sigir/MacAvaneySCGTF19, DBLP:conf/acl/GharebaghGF20, DBLP:conf/bionlp/KondadadiMNM21, DBLP:conf/bionlp/DaiWLZ21}, and progress notes \citep{DBLP:journals/artmed/MoenPHAPSS16, liang2019novel, DBLP:conf/naacl/AdamsAKZE21, DBLP:conf/coling/GaoDMXCA22}. In contrast, some studies have focused on dialogues \citep{yim2021towards,manas2021knowledge, DBLP:conf/emnlp/ZhangNGJHSG21}. Recently, \citet{DBLP:conf/lrec/GaoDMTLCA22} proposed the task of ``progress note understanding'', where the goal is to generate problem lists given the assessment sections of a progress note. They further explored the performance of \texttt{T5}~\citep{DBLP:journals/jmlr/RaffelSRLNMZLL20}, \texttt{BART}~\citep{DBLP:conf/bionlp/KondadadiMNM21} based on pre-training tasks with masked healthcare concepts \citep{DBLP:conf/coling/GaoDMXCA22}. To draw further attention to this task, The BioNLP 2023 Shared Task 1A \citep{gaobionlp} invited external participants to develop approaches to advance the state-of-the-art on the proposed task. 

\begin{figure*}[htbp]
    \centering
    \includegraphics[width=\linewidth]{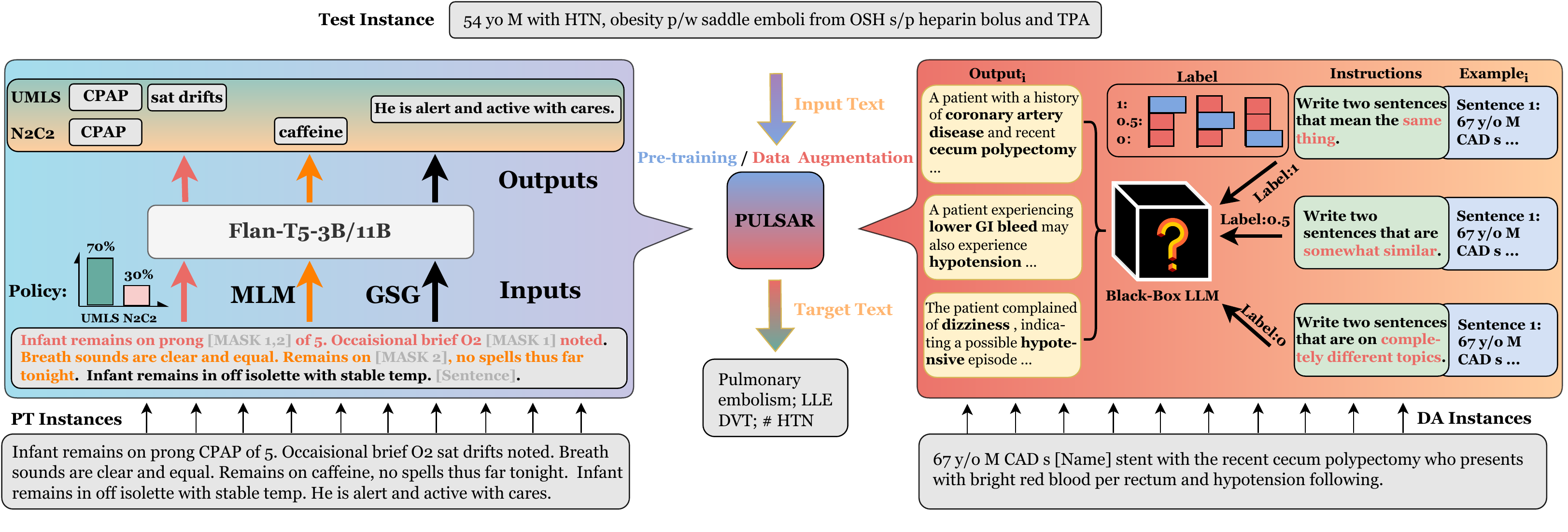}
    \caption{Overview of PULSAR. The left component represents the pre-training process with three different mask policies depicted in different colours. Both Gap Sentences Generation (GSG) and Masked Language Modelling (MLM)  are applied simultaneously to this example as pre-training objectives. The right component shows the workflow for data augmentation where the three labels $\left\{1, 0.5, 0\right\}$ represent \textsc{same thing}, \textsc{somewhat similar} and \textsc{completely different topics},  respectively. \textsc{PT Instances} and \textsc{DA Instances} stand for \textsc{Pre-training Instances} and \textsc{Data Augmentation Instances}, respectively.}
    \label{PULSARpipeline}
\end{figure*}

The main contribution of this work is a novel framework for data augmentation and summarisation of diagnoses/problems. In our approach, first, we optimise a domain-specific Language Model (LM) using a combination of different pre-training objectives, depicted in Figure~\ref{PULSARpipeline}; this model significantly outperforms the state-of-the-art, even when optimised on a limited number of manually annotated progress notes. Second, we instruct Large Language Models (LLMs) to generate synthetic data, in order to reduce the reliance on large, high-quality annotated datasets. Finally, we use the generated data to fine-tune the domain-specific LM on the task of problem list generation, given appropriate progress note sections. Our approach ranked second among all submissions to the shared task without additional annotated data. The results of our evaluation suggest that our pre-training objectives are aligned with the downstream task of summarisation and can significantly improve performance.

\section{Methodology}

Figure \ref{PULSARpipeline} shows the two components of our framework: first we pre-train an encoder-decoder model on MIMIC-III progress notes \citep{johnson2016mimic} using three different concept masking pre-training objectives. Then we employ data augmentation when fine-tuning our model for the summarisation task.

\subsection{Pre-training Model}

The items on the problem list are not necessarily directly extracted from the original progress notes and hence we cast the problem as abstractive summarisation. Drawing inspiration from  \texttt{PEGASUS} \citep{DBLP:conf/icml/ZhangZSL20}, we used an objective which closely resembles the abstractive summarisation objective, to gain better and faster fine-tuning performance. 

\begin{figure}[b!]
    \centering
    \includegraphics[width=\columnwidth]{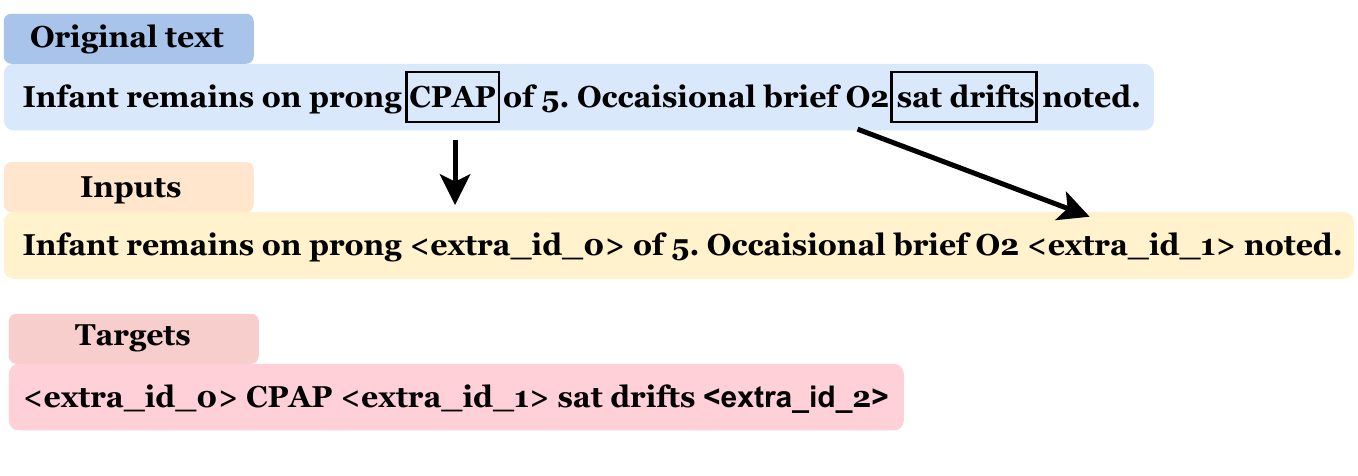}
    \caption{Our pre-training objective. The terms "CPAP" and "sat drifts" are identified by the NER models and replaced by a unique sentinel token respectively. The objective is to predict these masked-out spans.}
    \label{t5mlm}
\end{figure}

Following the success obtained through masking words and contiguous spans \citep{DBLP:journals/tacl/JoshiCLWZL20, DBLP:journals/jmlr/RaffelSRLNMZLL20}, we propose to select and mask text spans or whole sentences from input documents. We concatenate these ``gap text spans (sentences)'' into a pseudo-summary. Gap text spans were selected by the QuickUMLS entity linking \citep{soldaini2016quickumls} and an NER model trained on the i2b2-2010 challenge \citep{DBLP:journals/jamia/UzunerSSD11}. Similar to the T5 pre-training procedure \citep{DBLP:journals/jmlr/RaffelSRLNMZLL20}, these text spans were replaced by ``sentinel'' mask tokens $<extra \underline{~}id\underline{~}i>$ to inform the model that input was masked. Here, $i$ indicates the number of the mask (from left to right). The output sequence thus consists of the dropped-out text spans, delimited by the sentinel token between terms and the last $<extra \underline{~}id\underline{~}i>$  input representing the end of the output. Figure \ref{t5mlm} illustrates our pre-training objective. 

Specifically, 
we considered three masking policies in our pre-training objective. For each sentence, When both tools identified entities, we selected UMLS terms with the probability of $0.7$ and i2b2 terms with the probability of $0.3$. When only one tool identified entities, these entities were selected. Finally, when no entities were identified, the entire sentence was masked with a probability of $0.15$. In order to provide the model with the necessary medical knowledge and reduce domain barriers \citep{DBLP:journals/jksucis/PandeyPMR22}, we leverage all progress notes from MIMIC-III \citep{johnson2016mimic} to train \texttt{Flan-T5} \citep{DBLP:journals/corr/abs-2210-11416} on this objective. The processed pre-training corpus had 2.08m rows of data, with 2.2k containing no UMLS terms, 23k containing no i2b2 entities, and 797 where neither tool recognised any entities.

\subsection{Data Augmentation (DA)}

\begin{table}
    \centering
    \scalebox{0.7}{
    \begin{tabular}{lcc}
         \toprule
         {} & Dev set & Test set\\         
         \cmidrule(r){2-2} \cmidrule(r){3-3}
         Approach (Setting) & R-1/R-2/R-L & R-F1/R-P/R-R \\
         \midrule
         \emph{${\rm PULSAR_{3B}}(DA)$} & \colorbox{purpleforus}{\textbf{36.27}/\textbf{16.78}/\textbf{33.83}} & \colorbox{yellowdark}{30.48/38.02/29.72}\\
         \emph{${\rm PULSAR_{11B}}(DA)$}  & \colorbox{danshuhong}{35.92/15.87/33.14} & \colorbox{purpleforus}{\textbf{31.15}/40.93/28.73} \\
         \emph{${\rm PULSAR_{3B}}$}  & \colorbox{orangelight}{33.60/13.70/31.32} & \colorbox{danshuhong}{31.14/\textbf{44.30}/27.18}\\
         \emph{${\rm PULSAR_{11B}}$}  & \colorbox{yellowdark}{33.38/13.14/30.63} & \colorbox{bluedark}{30.34/42.68/27.12}\\
         \emph{${\rm FlanT5_{11B}}(DA)$}  & \colorbox{yalv}{32.57/13.07/29.95} & -\\
         \emph{${\rm FlanT5_{11B}}$}  &  \colorbox{bluedark}{31.24/11.42/28.25} & 30.06/40.61/27.25\\
         \emph{${\rm FlanT5_{3B}}(DA)$}  & 29.46/09.85/26.15 &\colorbox{yalv}{30.47/38.01/29.72}\\
         \emph{${\rm ClinicalT5_{LARGE}}(DA)$}  & 28.60/11.13/26.11 & 25.43/25.67/\textbf{32.05}\\
         \emph{${\rm FlanT5_{3B}}$}  & 28.90/08.93/25.26 & \colorbox{orangelight}{30.60/41.09/28.58}\\
         \emph{${\rm PULSAR_{3B}}(-A)$}  & 27.70/10.60/24.34 & 28.29/38.24/26.54\\
         \hdashline
         \emph{$ \rm ClinicalT5_{LARGE}$}  & 31.09/12.85/28.15 & 19.92/18.93/28.89 \\
         \emph{$\rm FlanT5_{LARGE}$} & 29.86/10.19/27.08 & - \\
         \bottomrule
    \end{tabular}
    }
    \caption{Performance of evaluated models on the development set  measured in terms of Rouge-1/2/LCS, and on the test set measured in terms of Rouge-F1/Precision/Recall, respectively. The composition of the input content is \textsc{Assessment + Subject + Object}, except where only the \textsc{Assessment} section of the input was used, indicated by \textsc{-A}. \textsc{DA} means that data augmentation was employed. The Rouge-L score on the development set was used for official ranking. Colours (i.e. \colorbox{purpleforus}{1st}, \colorbox{danshuhong}{2nd},\colorbox{orangelight}{3rd},\colorbox{yellowdark}{4th}, \colorbox{yalv}{5th},\colorbox{bluedark}{6th}) indicate the highest to lowest performance.
    }
    \label{rouge_result}
\end{table}

The lack of high-quality annotated data is a bottleneck that inhibits supervised learning methods in the healthcare field. For example, BioNLP Task 1A \citep{gaobionlp} has only 764 annotated training examples. Therefore, we rely on data augmentation techniques to obtain more training samples. Specifically, we propose a novel healthcare data generation (DG) framework based on DINO~\citep{DBLP:conf/emnlp/SchickS21a, li2023you}, which exploits the generative abilities of LLMs by relying on instruction following rather than model training. 
Our instructions to the LLMs include task-specific descriptions (i.e., ``\emph{Write two sentences that mean the same thing but keep these two healthcare terms $[Term 1],[Term 2]$}. Sentence 1: $[Source]$ Sentence 2: '') to make the model generate a paraphrase of $[Source]$, which is selected from  the annotated training data. The instructions to keep terms aim to keep relevant terms from $[Source]$ which also appear in the problem list (i.e. the output). In addition, we might expect that the text generated by the LLM would only fit well into the corresponding instruction but would not be applicable as a reasonable output for other instructions. For example, in Figure \ref{PULSARpipeline} (i.e. label $\left\{1\right\}$ is the expected label in blue and label$\left\{0\right\}$ is the count label in red), it is expected that the generated text should have the same meaning as $[Source]$, but at the same time not have a completely different meaning from $[Source]$. Following previous work, we employ the self-debiasing \citep{DBLP:journals/tacl/SchickUS21} algorithm to achieve this objective, i.e. when predicting the next token, not only the probability of the corresponding label is considered, but also the counter label is taken into account. We then use BERTScore \citep{DBLP:conf/iclr/ZhangKWWA20} and BLEURT \citep{DBLP:conf/acl/SellamDP20} to assess the similarity between each generated sample and the source, removing 85\% of the lowest scoring generated sentences. The backbone of the framework can be any generative LLM, such as \texttt{GPT3.5}\footnotemark[4], \texttt{GPT3} \citep{DBLP:conf/nips/BrownMRSKDNSSAA20} and \texttt{GPT2} \citep{radford2019language} series models. Limited by the data use agreement, we used \texttt{BioMedLM} \citep{bolton_hall_yasunaga_lee_manning_liang}, an open-source GPT-style model pre-trained on the biomedical abstracts and papers, 
\footnotetext[4]{chat.openai.com}
\footnote[5]{
The official test set result for PULSAR-11B was fine-tuned after the 0.33 pre-training epoch.}.

\subsection{Implementation Details}

\begin{table}
    \centering
    \scalebox{0.93}{
    \begin{tabular}{lc}
         \toprule
         Approach(MaxLen) & R-1/R-2/R-L \\
         \midrule
         \textbf{Baselines} & {}\\
         \emph{${\rm T5_{LARGE}}(512)$}  & 29.901/10.81/28.21 \\
         \emph{${\rm FlanT5_{BASE}}(512)$}  & \colorbox{lightpink}{27.16/8.9435/24.90}  \\
         \emph{${\rm ClinicalT5_{SCRATCH}}(512)$}  & 26.68/9.51/23.94 \\
         \emph{${\rm T5_{BASE}}(512)$}  & 25.07/7.72/23.36  \\
         \emph{${\rm FlanT5_{BASE}}(1024)$} & \colorbox{lightpink}{25.51/7.96/23.07}  \\
         \emph{${\rm ClinicalT5_{BASE}}(512)$}   & \colorbox{lightblue}{22.27/7.61/20.49} \\
         \emph{${\rm PEGASUS_{XSUM}}(512)$}  & 22.39/6.86/20.36  \\
         \emph{${\rm ClinicalT5_{BASE}}(1024)$}  & \colorbox{lightblue}{21.13/7.19/19.55}  \\
         \emph{${\rm ClinicalT5_{SCI}}(512)$}  & 14.12/4.61/13.22  \\
         \bottomrule
    \end{tabular}
    }
    \caption{Performance of baseline models on the development measured in terms of Rouge-1/2/LCS. The composition of the input content is \textsc{Assessment + Subject + Object}. The same colour represents the same model with different input lengths.
    }
    \label{baseline_result}    
\end{table}
\textbf{Pre-training}: We choose \texttt{FlanT5-3B} and \texttt{FlanT5-11B} \citep{DBLP:journals/corr/abs-2210-11416} as our LM. \texttt{PULSAR-3B} and \texttt{PULSAR-11B} are pre-trained on two NVIDIA Tesla A100 80GB GPUs and four NVIDIA Tesla A100 80GB GPUs for 1 epoch respectively\footnotemark[5].
During the pre-training, we rely on Fully Sharded Data Parallel (FSDP) with CPU offloading \citep{baines2021fairscale} to fit LLMs into GPU memory.

\textbf{Data Augmentation}: We employ \texttt{BioMedLM} \citep{radford2019language} as the data augmentation model with default settings, setting maximum output length to $40$. Finally, the generated data are matched with the corresponding summaries, subjective and objective to create a training set of $1$k instances. The DA model \citep{DBLP:conf/emnlp/SchickS21a} is run on a single NVIDIA Tesla V100 32G GPU, with each run taking up to twelve hours. Example templates and the full dataset description can be seen in Appendix~A.

\section{Experimental Setup}

\textbf{Baselines}: 
We have chosen to adapt T5-base as one of our baselines, similar to the approach taken by \citet{DBLP:conf/coling/GaoDMXCA22}. Additionally, we have incorporated various state-of-the-art models such as \texttt{FlanT5} \citep{DBLP:journals/corr/abs-2210-11416}, \texttt{ClinicalT5} \citep{lehman2023clinical} and \texttt{PEGASUS} \citep{DBLP:conf/icml/ZhangZSL20}. Whereas \texttt{FlanT5} is an enhanced version of T5 that has been finetuned in a mixture of tasks \cite{DBLP:journals/corr/abs-2210-11416} and \texttt{ClinicalT5} pre-trained on MIMIC-III \citep{johnson2016mimic}. \texttt{PEGASUS} is an abstractive summarisation model with Gap Sentences Generation  and  Masked Language Model \citep{DBLP:conf/naacl/DevlinCLT19} as pre-train tasks.

\textbf{Evaluation metrics}: We calculate ROUGE~\citep{lin2004rouge} scores on the test set, by comparing the generated summaries to their corresponding references, averaging for all generation samples. For all experiments, the data set was divided into a ``train'' and a ``dev'' set with a ratio of 8:2 for training and evaluation, respectively. The results are presented in Table~\ref{rouge_result}, left column, and Table \ref{baseline_result}.  Table~\ref{rouge_result}, right column, shows the performance of the models on the official withheld test set. In this case, both train and dev splits were used for training.

\section{Results and Analysis}

\textbf{Pre-training helps}: Both Table~\ref{rouge_result} and Table~\ref{baseline_result} demonstrate that the pre-training objective improves task performance (compare 3B and 11B \texttt{PULSAR} to corresponding \texttt{FlanT5} models). The best performance of \texttt{PULSAR} was 3.1 points higher than the \texttt{FlanT5-11B} on the development set as the training set and 11.2 points higher than \texttt{ClinicalT5-large} on the official test set. 
The small difference in performance between \texttt{PULSAR-11B} and \texttt{PULSAR-3B} is primarily because the former has only completed 1/3 of the first pre-training epoch, potentially resulting in a lack of relevant medical knowledge and familiarity with downstream task patterns.

\textbf{Data augmentation is effective when the data distribution is consistent; It is significantly more helpful for small models when on a random data distribution}: Table \ref{rouge_result} shows that, data augmentation improves performance (3 point on average, compared to not using DA). This shows that the proposed DA approach can effectively alleviate the lack of annotated healthcare data, when the distribution of training and test set data is consistent. From Table~\ref{rouge_result}, it becomes evident that smaller models (\texttt{ClinicalT5-large}) can improve by up to 6 points with the help of data augmentation, but the effect diminishes with model size as it increases max to 2.5 on LLMs. The potential reason is that the test set for the sharing task differs significantly from the training set, in the vary of length of the summary. 

\textbf{The model is capable of discriminating irrelevant information, but longer input lengths may result in decreased performance}: We conducted ablation experiments on \texttt{PULSAR-3B} to verify the impact of the input text type. In contrast to \citet{DBLP:conf/lrec/GaoDMTLCA22}'s findings on the small model, the results (\texttt{PULSAR-3B} vs. \texttt{PULSAR-3B-A}) in Table~\ref{rouge_result} show that if the input is \textsc{Assessment + Subjective + Objective}, the model performs better (by 2.9 points on the official test set and by 7 points on the development set) compared with only using \textsc{Assessment} as input. This indicates that while most of the relevant information can be inferred from the \textsc{Assessment} section alone, additional input can be beneficial.
However, increasing the input length appears to not be useful: Table~\ref{baseline_result} shows that models trained with longer input lengths (1024 tokens) 
do not improve over models that were trained on 512-token-long input.

\section{Conclusion}

This paper contributed to the development of summarising patients’ problems. Firstly, we proposed a novel task-specific pre-training LLM objective. Compared with other submissions, we rank 2nd at the official shared task without using additional manually annotated training samples. Secondly, we propose a new data augmentation framework and demonstrate its effectiveness in the healthcare domain. In the future, we will explore the applicability of our approach to other domain-specific generative tasks and conduct a deeper analysis of factors that contribute to overall model performance. 

\section*{Limitations}

The proposed model is computationally demanding. Recent work on parameter-efficient fine-tuning methods, such as LoRA \citep{DBLP:conf/iclr/HuSWALWWC22}, suggests that they can  significantly reduce the number of trainable parameters at a minimal performance cost, which may help further democratise the development of domain- and task-specific models. In addition, as we continued to pretrain, to obtain the \texttt{PULSAR} models, their tokenizer was inherited from corresponding \texttt{Flan-T5} model. Thus it does not contain domain-specific terminology, which may be a limitation in terms of representation density (i.e. frequent clinical terms may be split in multiple rare sub-tokens).

\section*{Ethics Statement}

For the present work, we used an existing anonymised dataset from BioNLP 2023 Shared Task 1A without any data protection issues. In addition, data augmentation only uses an open-source, off-line model which is not offensive to the data user agreement that is shared with a third party.

\section*{Acknowledgements}
We thank the anonymous reviewers from the BioNLP 2023 Shared Task for their valuable feedback. We would also like to acknowledge the use of the Computational Shared Facility at The University of Manchester. 

\bibliography{anthology,custom}
\bibliographystyle{acl_natbib}

\appendix

\section{Example Appendix}
\label{sec:appendix}
Example of data augmentation input and output

\begin{figure}[htbp]
    \centering
    \includegraphics[width=\columnwidth]{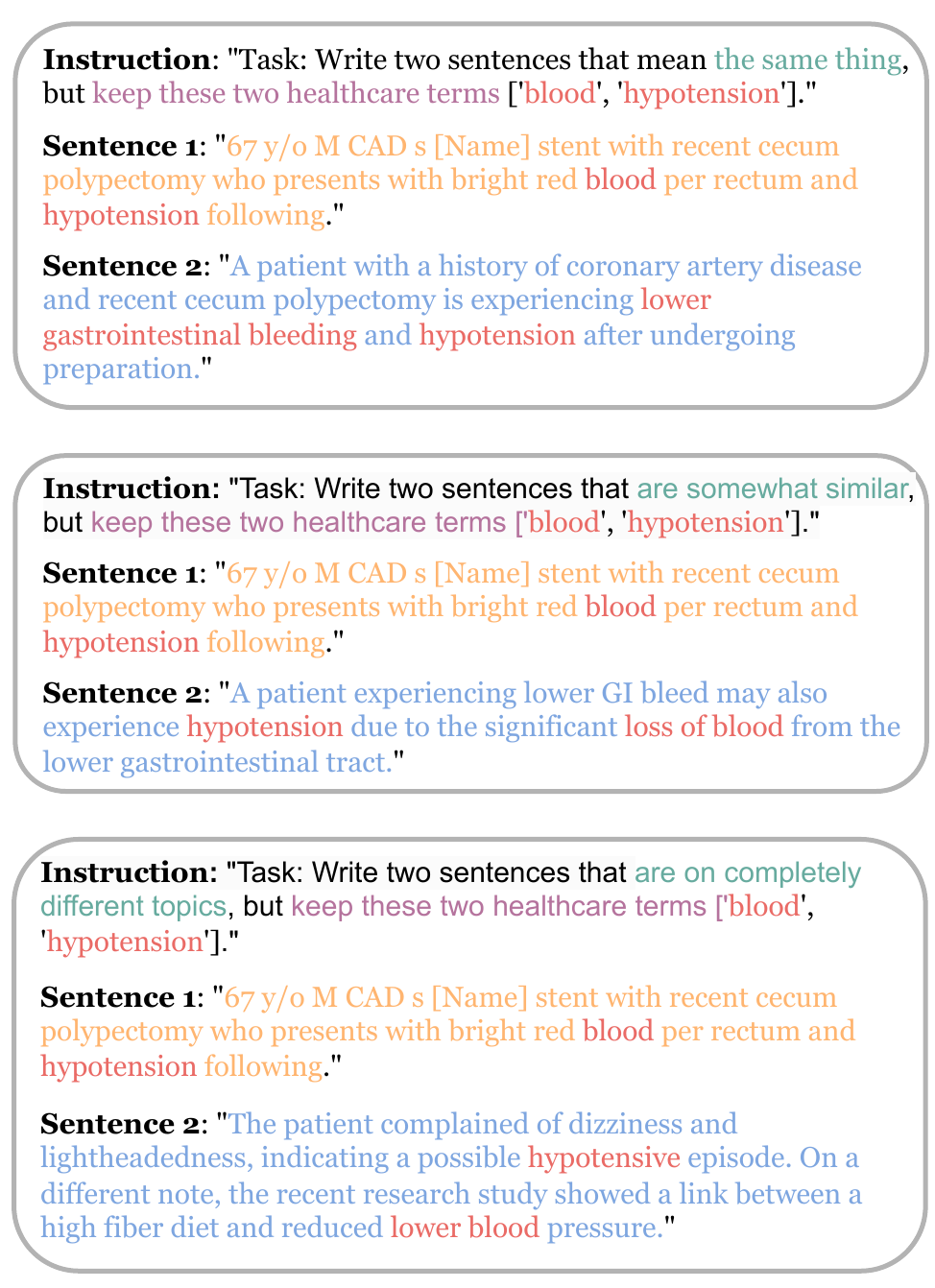}
    \caption{Continuation text generated by prompted learning data augmented methods with three different template descriptions. We chose to give input sentence 1 and generate only sentence 2, which helps to generate sentence similarity datasets.}
    \label{medicalDG}
\end{figure}

\end{document}